\documentclass[]{article}

\PassOptionsToPackage{numbers,sort&compress}{natbib}  
\usepackage[preprint]{neurips_2022}
\usepackage[utf8]{inputenc} 
\usepackage[T1]{fontenc}    
\usepackage{hyperref}       
\usepackage{url}            
\usepackage{booktabs}       
\usepackage{amsfonts}       
\usepackage{amssymb}
\usepackage{pifont}
\newcommand{\cmark}{\ding{51}}%
\newcommand{\xmark}{\ding{55}}%

\usepackage{nicefrac}       
\usepackage{microtype}      
\usepackage{xcolor}         
\usepackage{graphicx}       
\usepackage[normalem]{ulem} 
\useunder{\uline}{\ul}{}    
\usepackage{amsmath}        
\usepackage{xspace}
\usepackage{comment}
\usepackage{subfigure}      
\usepackage{svg}            
\usepackage{color}
\usepackage{marvosym}

\newcommand{\ours}[0]{FreeGEM\xspace}

\title{Parameter-free Dynamic Graph Embedding \\ for Link Prediction}

\author{
Jiahao Liu\textsuperscript{1,2}, Dongsheng Li\textsuperscript{3}, Hansu Gu\textsuperscript{4\Letter}, Tun Lu\textsuperscript{1,2\Letter}, Peng Zhang\textsuperscript{1,2}, Ning Gu\textsuperscript{1,2} \\
\textsuperscript{1}School of Computer Science, Fudan University, Shanghai, China \\
\textsuperscript{2}Shanghai Key Laboratory of Data Science, Fudan University, Shanghai, China \\
\textsuperscript{3}Microsoft Research Asia, Shanghai, China \quad \textsuperscript{4}Seattle, United States \\
\texttt{jiahaoliu21@m.fudan.edu.cn, dongsli@microsoft.com, hansug@acm.org} \\
\texttt{\{lutun, zhangpeng\_, ninggu\}@fudan.edu.cn}
}

\begin{document}

\maketitle

\begin{abstract}
Dynamic interaction graphs have been widely adopted to model the evolution of user-item interactions over time.
There are two crucial factors when modelling user preferences for link prediction in dynamic interaction graphs: 1) collaborative relationship among users and 2) user personalized interaction patterns.
Existing methods often implicitly consider these two factors together, which may lead to noisy user modelling when the two factors diverge. In addition, they usually require time-consuming parameter learning with back-propagation, which is prohibitive for real-time user preference modelling.
To this end, this paper proposes \ours, a parameter-free dynamic graph embedding method for link prediction.
Firstly, to take advantage of the collaborative relationships, we propose an incremental graph embedding engine to obtain user/item embeddings, which is an Online-Monitor-Offline architecture consisting of an Online module to approximately embed users/items over time, a Monitor module to estimate the approximation error in real time and an Offline module to calibrate the user/item embeddings when the online approximation errors exceed a threshold.
{Meanwhile, we integrate attribute information into the model, which enables FreeGEM to better model users belonging to some under represented groups.}
Secondly, we design a personalized dynamic interaction pattern modeller, which combines dynamic time decay with attention mechanism to model user short-term interests.
Experimental results on two link prediction tasks show that \ours can outperform the state-of-the-art methods in accuracy while achieving over 36X improvement in efficiency.
{All code and datasets can be found in \url{https://github.com/FudanCISL/FreeGEM}.}
\end{abstract}

\section{Introduction}
Dynamic interaction graphs have been adopted in a wide range of link prediction tasks to model the evolution of user-item interactions over time~\cite{kazemi2020representation,kumar2019predicting,zhang2021cope}.
The evolution of a dynamic interaction graph can be reflected by its historical interaction (edge) sequence, in which two crucial factors should be explicitly considered for link prediction tasks:
1) collaborative relationship among users, i.e., users with similar historical interactions will interact with similar items in the future, which is also the basic assumption of collaborative filtering~\cite{goldberg1992using,sarwar2001item,Li2017nips}; and 
2) user personalized dynamic interaction patterns, i.e., users could have unique short-term interaction patterns, which are not shared among like-minded users but can only be reflected by their own interaction sequences. 

Many existing methods learn user preferences from dynamic interaction graphs using temporal point process (TPP)~\cite{trivedi2017know,zuo2018embedding,wang2016coevolutionary}, recurrent neural network (RNN)~\cite{kumar2019predicting,wu2017recurrent,beutel2018latent} and graph neural network (GNN)~\cite{chang2020continuous,shen2021powerful,zhang2021cope}, etc., in which the following key challenges arise.
Firstly, these methods do not consider the above two key factors separately, so that collaborative relationship may bring noises to personalized patterns when they diverge, and vice versa. 
Secondly, the above methods often require time-consuming parameter learning with back-propagation. In dynamic interaction graphs, the model training should follow chronological order of the interactions to capture the temporal dynamics, which raises efficiency issue even for applications with moderate number of interactions.

\begin{figure}[t]
  \centering
  \includegraphics[width=1.0\linewidth]{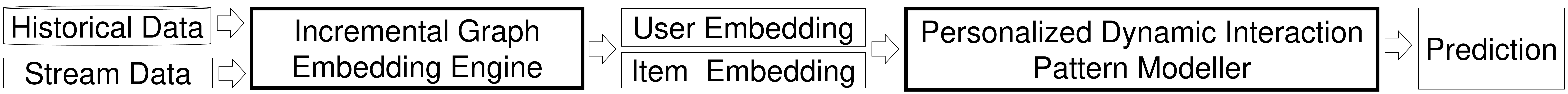}
  \caption{The high-level design of the proposed \ours method.}
  \label{fig.top}
\end{figure}

In this paper, we propose a Parameter-\textbf{Free} Dynamic \textbf{G}raph \textbf{EM}bedding (\ours) method for link prediction. \textit{Here, parameter-free means that we do not incorporate any parameters which need to be learned via back-propagation.} 
As shown in Figure~\ref{fig.top}, \ours consists of two key components: incremental graph embedding engine and personalized dynamic interaction pattern modeller.
The incremental graph embedding engine takes both historical data and online stream data as input and  generates user/item embeddings as the output, which exploits the collaborative relationship among users.
Its core innovation is the proposed Online-Monitor-Offline architecture, which can achieve online embedding updates by solving closed-form solutions in real time and keep the approximation errors caused by online singular value decomposition (SVD)~\cite{brand2006fast} within any predefined threshold.
In the Offline step, we propose a frequency-aware preference matrix reconstruction method to alleviate the oversmoothing problem and an attribute-integrated SVD to alleviate the cold-start issue.
{Surprisingly, the integration of attribute information enables \ours to better model users belonging to under represented groups.}
In the Online step, we convert the offline truncated SVD into an online SVD to generate embeddings in real time.
In the Monitor module, we estimate the online approximation error in real time by analyzing the relationship between approximation error and the update of online algorithm.
The personalized dynamic interaction pattern modeller is also a parameter-free component, which combines the dynamic time decay with attention mechanism to model user short-term interests.
It takes the user and item embeddings as input and outputs the prediction results. 
Specifically, it selectively forgets the early interactions through dynamic time decay and hence focuses on more recent interactions for prediction.
Then, it leverages the attention mechanism to capture user personalized dynamic interaction patterns over the ``decayed'' item embedding sequences.
Experimental results on two link prediction tasks (future item recommendation and next interaction prediction) show that \ours can substantially outperform the state-of-the-art link prediction methods in accuracy while achieving over 36X improvement in computational efficiency. Besides, our empirical studies also confirm that \ours can alleviate the cold-start issue and achieve high robustness on very sparse data.

\section{Related Work}
Link prediction methods on dynamic interaction graphs are mostly based on TPP, RNN and GNN, etc., which need time-consuming training process. To the best of our knowledge, the proposed \ours is the only parameter-free method in this task, with high accuracy and efficiency.

\paragraph{TPP-based methods}
Know-Evolve~\cite{trivedi2017know} and HTNE~\cite{zuo2018embedding} model interactions as multivariate point processes and Hawkes processes, respectively.
Wang et al.~\cite{wang2016coevolutionary} propose a co-evolutionary process to model the co-evolving nature of users and items.
Shchur et al.~\cite{shchur2019intensity} propose to directly model the conditional distribution of inter-event times.
DSPP~\cite{cao2021deep} incorporates topology and long-term dependencies into the intensity function.

\paragraph{RNN-based methods}
RRN~\cite{wu2017recurrent} models user and item interaction sequences with separate RNNs.
Time-LSTM~\cite{zhu2017next} proposes time gates to represent the time intervals.
LatentCross~\cite{beutel2018latent} incorporates contextual data into embeddings.
DeepCoevolve~\cite{dai2016deep} and JODIE~\cite{kumar2019predicting} generate node embeddings using two intertwined RNNs. JODIE~\cite{kumar2019predicting} can also estimate user embeddings trajectories.
DeePRed~\cite{kefato2021dynamic} employs non-recursive mutual RNNs to model interactions.

\paragraph{GNN-based methods}
TDIG-MPNN~\cite{chang2020continuous} captures the global and local information on the graph.
DGCF~\cite{li2020dynamic} uses three update mechanisms to update users and items.
SDGNN~\cite{tian2021streaming} takes the state changes of neighbor nodes into account.
MRATE~\cite{chen2021multi} combines different relations to realize multi-relation awareness.
OGN~\cite{kang2021online} updates nodes in an online fashion with a constant memory cost.
TCL~\cite{wang2021tcl} proposes a graph-topology-aware transformer to learn the representations of nodes.
CoPE~\cite{zhang2021cope} uses an ordinary differential equation-based GNN to model the evolution of network.
TGL~\cite{zhou2022tgl} proposes a unified framework for large-scale temporal GNN training.
MetaDyGNN~\cite{yang2022few} proposes a model based on
a meta-learning framework for few-shot link prediction in dynamic networks.
TREND~\cite{wen2022trend} uses Hawkes process-based GNN for temporal graph representation learning.

\section{Incremental Graph Embedding Engine}
\begin{figure}[t]
  \centering
\includegraphics[width=0.8\linewidth]{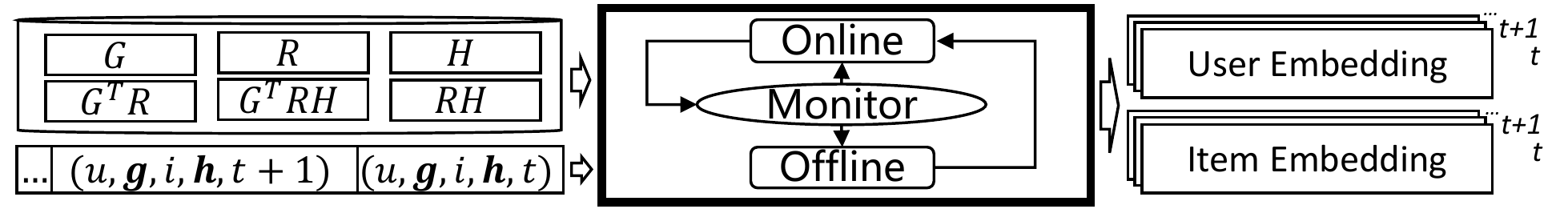}
    \caption{Illustration of the proposed incremental graph embedding engine.}
    \label{fig.engine}
\end{figure}

As shown in Figure~\ref{fig.engine}, the incremental graph embedding engine 1) takes the {historical} user-item interaction, user/item attribute {and online stream data} as the input, 2) embeds users and items via the novel Online-Monitor-Offline architecture, and 3) outputs the user and item embeddings in real time.

\paragraph{Online-Monitor-Offline Architecture}
The Offline module first decomposes the historical data matrices by offline SVD to obtain user and item embeddings.
To capture real-time information from the interaction stream, the model needs to efficiently update the corresponding embeddings when new interaction occurs.
Thus, we propose the Online module, which uses online SVD to approximately update user and item embeddings, to meet the real-time requirements.
However, since the approximation error of online SVD accumulates over time, 
we propose a Monitor module to estimate the accumulated approximation error of the Online module in real time.
When the accumulated error exceeds a threshold, we restart the Offline module to calibrate the user and item embeddings.
Otherwise, we continue to execute the Online module for real-time user/item embedding.

\subsection{Offline Module}
The input data of \ours includes six matrices as defined in Table~\ref{table.embedding}, in which $R\in\mathbb{R}^{m\times n}$ is user-item interaction matrix, $G\in\mathbb{R}^{m\times p}$ is user attribute matrix and $H\in\mathbb{R}^{n\times q}$ is item attribute matrix, where $m$/$n$ is the number of users/items, $q$/$p$ is the dimension of user/item attribute vector and $(R)_{ij}$ indicates the number of interactions between user $i$ and item $j$.
Each stream data sample can be represented as a 5-tuple $(u,\mathbf{g},i,\mathbf{h},t)$, where $u$, $i$, $t$ are user id, item id and timestamp respectively, and $\mathbf{g}\in\mathbb{R}^{p}$ and $\mathbf{h}\in\mathbb{R}^{q}$ are user and item attribute vectors, respectively.

\subsubsection{Frequency-aware Preference Matrix Reconstruction}
\label{sec.frequency}
There are three steps to realize collaborative filtering~\cite{sarwar2000application}:
1) decompose the interaction matrix $R$ using truncated SVD to obtain $U=(\mathbf{u}_1,...,\mathbf{u}_k)\in\mathbb{R}^{m\times k}$, $S=diag(s_1,...,s_k)\in\mathbb{R}^{k\times k},s_1>...>s_k>0$ and $V=(\mathbf{v}_1,...,\mathbf{v}_k)\in\mathbb{R}^{n\times k}$;
2) the embeddings of users and items are $E_U=US^{1/2}$ and $E_I=VS^{1/2}$, respectively; and 
3) use $\hat{R}=E_UE_I^\top$ as the low rank approximation of $R$.
{Our method makes non-trivial improvements based on this, including frequency-aware preference matrix reconstruction in this section and attribute-integrated SVD in the Section \ref{sec.att}.}

The raw interaction matrix $R$ cannot perfectly reflect user preferences due to potential biases~\cite{steck2011item,steck2019markov}, so we propose to normalize $R$ as follows:  ${R'}=D_U^{-\alpha}RD_I^{-\alpha}$, where $\alpha>0$ is a hyperparameter, $D_U$ and $D_I$ are diagonal matrices, and $(D_U)_{ii}=\Sigma_{j=1}^nR_{ij}$ and $(D_I)_{jj}=\Sigma_{i=1}^mR_{ij}$.
Compared with $R$, 
${R'}$ can more accurately represent user preferences. Intuitively, the more users an item interacts with, the less it can reflect each user's preference, and vice versa. The normalization can alleviate the popular bias and help to achieve more accurate user representation.

By applying {offline} truncated SVD on ${R'}$, we have the approximation of ${R'}$ as: $\hat{{R'}}=\Sigma_{i=1}^ks_i\mathbf{u}_i\mathbf{v}_i^\top$, which only retains the low-frequency signals of ${R'}$, so it can be regarded as ${R'}$ passing through an ideal low-pass graph filter. Thus, it will suffer from the oversmoothing problem when $k$ is small.
To alleviate this problem, we introduce a hyperparameter $\gamma$ to control the ratio of high-frequency signals to low-frequency signals as follows: $E_U=US^{\gamma}$, $E_I=VS^{\gamma}$  ($\gamma<0.5$).
The original intensity of each frequency is $s_i$, which becomes {\small$s_i^{2\gamma}$} after applying the frequency control. The attenuation ratio is {\small${s_i^{{2\gamma}}}/{s_i}$}, which is a hyperbola about $s_i$.
The low-frequency signal corresponds to a larger $s_i$, which means that the attenuation of the low-frequency signal is stronger than that of the high-frequency signal, improving the proportion of the high-frequency signal in the reconstructed matrix $\hat{{R'}}$.
In summary, this is equivalent to reducing the proportion of low-frequency signals after ideal low-pass filtering through truncated SVD {(more
discussion can be found in the Appendix \ref{app.fre})}.
Finally, the predicted interaction matrix is obtained by inverse normalization as follows: $\hat{R}=D_U^{\alpha}\hat{{R'}}D_I^{\alpha}$.

\subsubsection{Attribute-integrated SVD}
\label{sec.att}
\begin{table}[]\scriptsize
\caption{Correspondence between the decomposed matrix and the constructed embedded matrix.}
\label{table.embedding}
\centering
\begin{tabular}{c|c|c|c}
\hline
Description of the decomposed matrix & Decomposed matrix & Constructed embedding (left) & Constructed embedding (right) \\ \hline
user-item & $R\in\mathbb{R}^{m\times n}$ & $E_U^{1}\in\mathbb{R}^{m\times k_{1}}$ & $E_I^{1}\in\mathbb{R}^{n\times k_{1}}$ \\
user attribute & $G\in\mathbb{R}^{m\times p}$ & $E_U^{2}\in\mathbb{R}^{m\times k_{2}}$ & $E_G^{2}\in\mathbb{R}^{p\times k_{2}}$ \\
item attribute & $H\in\mathbb{R}^{n\times q}$ & $E_I^{3}\in\mathbb{R}^{n\times k_{3}}$ & $E_H^{3}\in\mathbb{R}^{q\times k_{3}}$ \\
user\_attribute-item & $G^\top R\in\mathbb{R}^{p\times n}$ & $E_G^{4}\in\mathbb{R}^{p\times k_{4}}$ & $E_I^{4}\in\mathbb{R}^{n\times k_{4}}$ \\
user-item\_attribute & $RH\in\mathbb{R}^{m\times q}$ & $E_U^{5}\in\mathbb{R}^{m\times k_{5}}$ & $E_H^{5}\in\mathbb{R}^{q\times k_{5}}$ \\
user\_attribute-item\_attribute & $G^\top RH\in\mathbb{R}^{p\times q}$ & $E_G^{6}\in\mathbb{R}^{p\times k_{6}}$ & $E_H^{6}\in\mathbb{R}^{q\times k_{6}}$ \\ \hline
\end{tabular}
\end{table}

\begin{table}[]\scriptsize
\caption{Correspondence between path and embedding matrix.}
\label{table.path}
\centering
\begin{tabular}{c|l|l|l}
\hline
No. & Path & User embedding & Item embedding \\ \hline
1 & {user}-{item} & $E_{U_1}=E_U^1\in\mathbb{R}^{m\times k_1}$ & $E_{I_1}=E_I^1\in\mathbb{R}^{n\times k_1}$ \\
2 & {user}-{user\_attribute}-{item} & $E_{U_2}=E_U^2(E_G^2)^\top\in\mathbb{R}^{m\times p}$ & $E_{I_2}=E_I^4(E_G^4)^\top\in\mathbb{R}^{n\times p}$ \\
3 & {user}-{item\_attribute}-{item} & $E_{U_3}=E_U^5(E_H^5)^\top\in\mathbb{R}^{m\times q}$ & $E_{I_3}=E_I^3(E_H^3)^\top\in\mathbb{R}^{n\times q}$ \\
4 & {user}-{user\_attribute}-{item\_attribute}-{item} & $E_{U_4}=E_U^2(E_G^2)^\top E_G^6\in\mathbb{R}^{m\times k_6}$ & $E_{I_4}=E_I^3(E_H^3)^\top E_H^6\in\mathbb{R}^{n\times k_6}$ \\
5 & {user}-{item\_attribute}-{user\_attribute}-{item} & $E_{U_5}=E_U^5(E_H^5)^\top E_H^6\in\mathbb{R}^{m\times k_6}$ & $E_{I_5}=E_I^4(E_G^4)^\top E_G^6\in\mathbb{R}^{n\times k_6}$ \\
\hline
\end{tabular}
\end{table}

{In addition to frequency-aware preference matrix reconstruction, compared with the plain SVD, we also integrate attribute information to better model users and items.}
Integrating user and item attributes can help to improve the prediction accuracy and alleviate the cold-start issue in the recommender system~\cite{zeng2021knowledge,xia22www}.
$R$, $G$ and $H$ are three basic matrices, which respectively describe the co-occurrence relationship of {user}-{item}, {user}-{user\_attribute} and {item}-{item\_attribute}.
{To connect them,} we further obtain three derived matrices $G^\top RH\in\mathbb{R}^{p\times q}$, $G^\top R\in\mathbb{R}^{p\times n}$ and $RH\in\mathbb{R}^{m\times q}$, which describe the co-occurrence relationship of {user\_attribute}-{item\_attribute}, {user\_attribute}-{item} and {user}-{item\_attribute} respectively.
After decomposing and reconstructing the six matrices using the method described in Section~\ref{sec.frequency}, we can get $6\times 2=12$ embedding matrices, as shown in Table~\ref{table.embedding}, whose superscript indicate embedding space, corresponding to different dimensions.
As shown in Figure~\ref{fig.side}, we have four objects: {user}, {user attribute}, {item} and {item attribute}.
There are $C_4^2=6$ relationships between them, which are represented by {the} six matrices respectively.
The six edges represent six embedding spaces, and each object has a representation in three of them.
There are five paths between {user} and {item} without revisiting, each of which corresponds to a user embedding matrix and an item embedding matrix, as shown in Table~\ref{table.path}.
Except for the {user}-{item} path where {user} and {item} are directly connected, all other paths go through {user/item attribute} and thus integrate the user/item attributes in the embeddings.
Finally, we only concatenate the embeddings from the first three paths as the output of the Offline module, for the last two paths consider user/item attributes repeatedly, {and obtain user embedding and item embedding as follows}:
\begin{align}
E_U&=(\alpha_1E_{U_1})||(\alpha_2E_{U_2})||(\alpha_3E_{U_3})\in\mathbb{R}^{m\times (k_1+p+q)}.\label{eqn:embed_u}\\
E_I&=(\alpha_1E_{I_1})||(\alpha_2E_{I_2})||(\alpha_3E_{I_3})\in\mathbb{R}^{n\times (k_1+p+q)}.
\label{eqn:embed_i}
\end{align}
$\alpha_i$ is the hyperparameter that controls the weight of the $i$-th path, and $||$ represents concatenation.

\subsection{Online Module}
The only difference between Online module and Offline module is the way to obtain  truncated SVD, where the Offline module uses ordinary truncated SVD but the Online module uses online SVD~\cite{brand2006fast}. 
Online SVD~\cite{brand2006fast} provides an approximated method to calculate the truncated SVD of an updated matrix in linear time.
Let $R_t\in\mathbb{R}^{m\times n}$ be the interaction matrix at time $t$.
Assuming that we have finished the truncated SVD of $R_t$, and then at time $t+1$, user $u$ interacts with item $i$. Thus, we have $R_{t+1}=R_t+\mathbf{ui}^\top$, where $\mathbf{u}\in\mathbb{R}^m$ and $\mathbf{i}\in\mathbb{R}^n$ are one-hot vectors indicating user id and item id, respectively.
Therefore, we realize the incremental calculation of truncated SVD of $R_{t+1}$ given the truncated SVD of $R_{t}=USV^\top$, where $U\in\mathbb{R}^{m\times k}$, $S\in\mathbb{R}^{k\times k}$, $V\in\mathbb{R}^{n\times k}$ and $k=k_1+p+q$ as shown in Equation~\ref{eqn:embed_u} and Equation~\ref{eqn:embed_i}. The details are described as follows.

Firstly, we calculate:
\begin{equation}
\mathbf{m}=U^\top\mathbf{u},\ 
\mathbf{n}=V^\top\mathbf{i},\ 
\mathbf{p}=\mathbf{u}-U\mathbf{m},\ 
\mathbf{q}=\mathbf{i}-V\mathbf{n},\ 
P=||\mathbf{p}||^{-1}\mathbf{p},\ 
Q=||\mathbf{q}||^{-1}\mathbf{q}.
\end{equation}
Secondly, we calculate:
\begin{align}
K=\begin{bmatrix} S & \mathbf{0} \\ \mathbf{0} & 0 \end{bmatrix} + \begin{bmatrix} \mathbf{m} \\ ||\mathbf{p}|| \end{bmatrix}\begin{bmatrix} \mathbf{n} \\ ||\mathbf{q}|| \end{bmatrix}^\top.
\end{align}
Thirdly, we calculate the full SVD of $K$ and get $U_KS_KV_K^\top$ with $U_K\in\mathbb{R}^{(k+1)\times (k+1)}$, $V_K\in\mathbb{R}^{(k+1)\times (k+1)}$ and $S_K\in\mathbb{R}^{(k+1)\times (k+1)}$.
Then, we have the following result:
\begin{align}
R_{t+1}=R_t+\mathbf{ui}^\top\approx([U\ P]U_K)S_K([V\ Q]V_K))^\top.
\end{align}
Finally, we can use the first $k$ columns of $[U\  P]U_K$, $S_K$ and $[V\ Q]V_K$ as the estimate of the truncated SVD of $R_{t+1}$.

\subsection{Monitor Module}

\begin{figure}[t]
  \centering
  \begin{minipage}[t]{0.45\linewidth}
    \centering
    \includegraphics[width=1\linewidth]{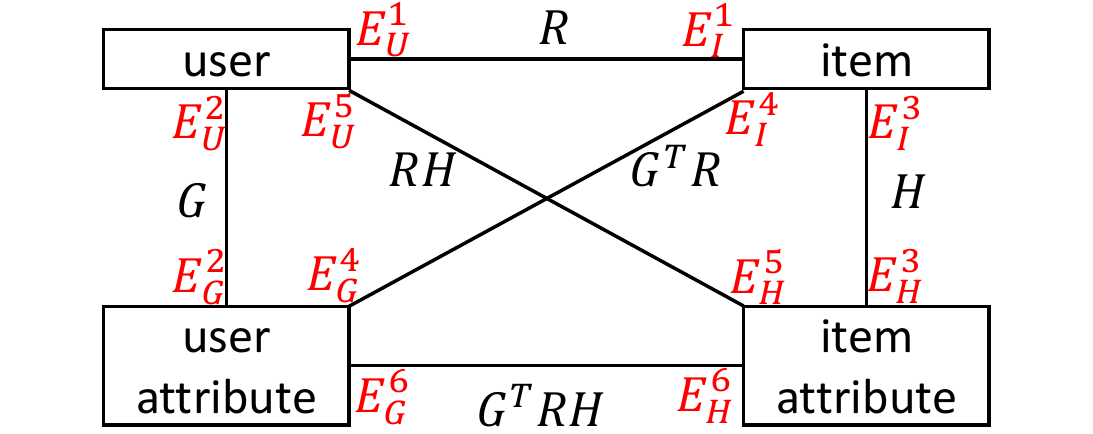}
    \caption{Attribute-integrated SVD.}
    \label{fig.side}
  \end{minipage}
  \quad
  \begin{minipage}[t]{0.5\linewidth}
    \centering
    \includegraphics[width=1\linewidth]{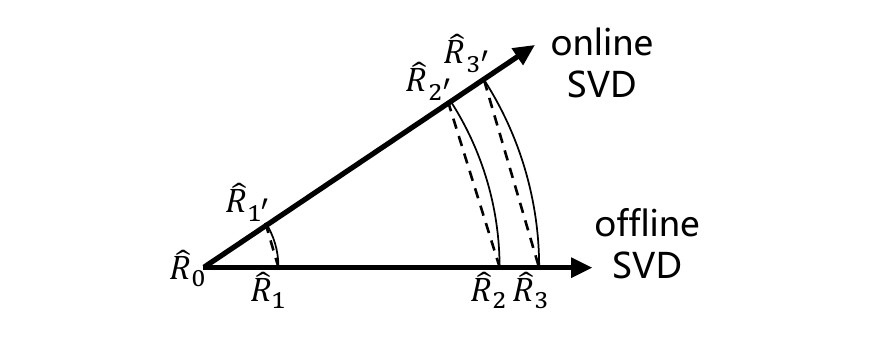}
    \caption{Motivating example of the Monitor.}
    \label{fig.monitor.motivation}
  \end{minipage}
\end{figure}

In the Online module, the approximation error will accumulate over time. To analyze the accumulated error, 
TIMERS~\cite{zhang2018timers} was proposed to calculate the lower bound of the approximation error of online SVD through matrix perturbation~\cite{stewart1990matrix}.
However, TIMERS depends on a time-consuming eigen-decomposition, so it can not monitor the error in real time but based on the granularity of time slice {for eigen-decomposition}.
Nevertheless, TIMERS provides two insights for the design of the Monitor:
1) the accumulated approximation error does not change uniformly with time and
2) there are no shared correlation patterns between the running time of online SVD, the number of new observations in the interaction matrices and the accumulated error among different datasets.

The timing of restart Offline module is very important because 
untimely restart will lead to excessive error and reduce the embedding quality and too frequent restart will lead to low efficiency.
There could be two restart heuristics: 1) restart after a certain time and 2) restart after a certain number of new interactions.
However, both heuristics are not optimal due to unawareness of the online approximation error, leading to potentially unnecessary or inaccurate updates.

The motivation of the Monitor is illustrated in Figure~\ref{fig.monitor.motivation}.
At time step 0, we initialize the offline truncated SVD of the user-item interaction matrix $\hat{R}_0$. 
At time step $i$, we can have matrices: 1) $\hat{R}_i$ if we use offline SVD and 2) $\hat{R}_{i'}$ if we use online SVD to approximate the matrix when new interaction occurs.
For ease of illustration, we use straight lines to represent the F-norm distance of the reconstructed matrices, and arrows to represent the evolution directions of the results.
The angle between the evolution directions of offline SVD and online SVD should be less than $\pi/3$ {(more discussion can be found in the Appendix \ref{app.pi})}, otherwise the results of online SVD is even worse than not updating at all.
Since there is no online approximation error in offline SVD, we take {$\hat{R}_{i}$, where $i=1,2,3$,} as the ground truth and {$\hat{R}_{i'}$, where $i=1,2,3$,} correspond to the results obtained by online SVD after each update.
The lengths of {$\hat{R}_0$$\hat{R}_1$ ($\hat{R}_0$$\hat{R}_{1'}$), $\hat{R}_1$$\hat{R}_2$ ($\hat{R}_{1'}$$\hat{R}_{2'}$) and $\hat{R}_2$$\hat{R}_3$ ($\hat{R}_{2'}$$\hat{R}_{3'}$)} are not equal, for different interactions have different impacts on the evolution of SVD~\cite{zhang2018timers}.
The dotted line represents the F-norm distance between the matrices reconstructed by offline SVD and online SVD, {$e=||\hat{R}_i-\hat{R}_{i'}||_F$}, i.e., the online approximation error.
The dotted line become longer over time, indicating that the error will accumulate with the occurrence of new interactions. 

We use \textit{distance} to specifically refer to the F-norm distance between the reconstructed matrix by online SVD at the current time step and the initial reconstructed matrix: {$d=||\hat{R}_{i'}-\hat{R}_{0}||_F$}.
Calculating $\hat{R}_1$, $\hat{R}_2$, $\hat{R}_3$ is time-consuming, so we hope to monitor the length of the dotted line without calculating it.
While calculating $\hat{R}_{1'}$, $\hat{R}_{2'}$, $\hat{R}_{3'}$ is efficient due to the linearity of online SVD and
it can be seen from the Figure~\ref{fig.monitor.motivation} that
there is a positive correlation between \textit{distance} and approximation error (empirically verified in the Appendix~\ref{ap:monitor}), so we can estimate the online approximation error though the \textit{distance}.
Every time the Online module is executed, the Monitor will calculate the \textit{distance} in real time.
When the \textit{distance} exceeds the predefined threshold, we believe the corresponding approximation error also reaches another threshold for restarting the Offline module.

Compared with the two simple heuristics, our Monitor estimates errors in a data-driven way with the following benefits: 1) the Monitor estimates the error according to the interaction events, avoiding the negative impacts caused by uneven interaction distributions in different time intervals; and
2) the Monitor estimates the error according to the positive correlation between the \textit{distance} and the error, avoiding the varying impacts of different interactions on the error estimation.
It should be noted that, compared with the two heuristics, Monitor only adds the process of calculating the \textit{distance}, which is very efficient and thus can achieve real-time monitoring.

\section{Personalized Dynamic Interaction Pattern Modeller}
\begin{figure}[t]
  \centering
  \includegraphics[width=1\linewidth]{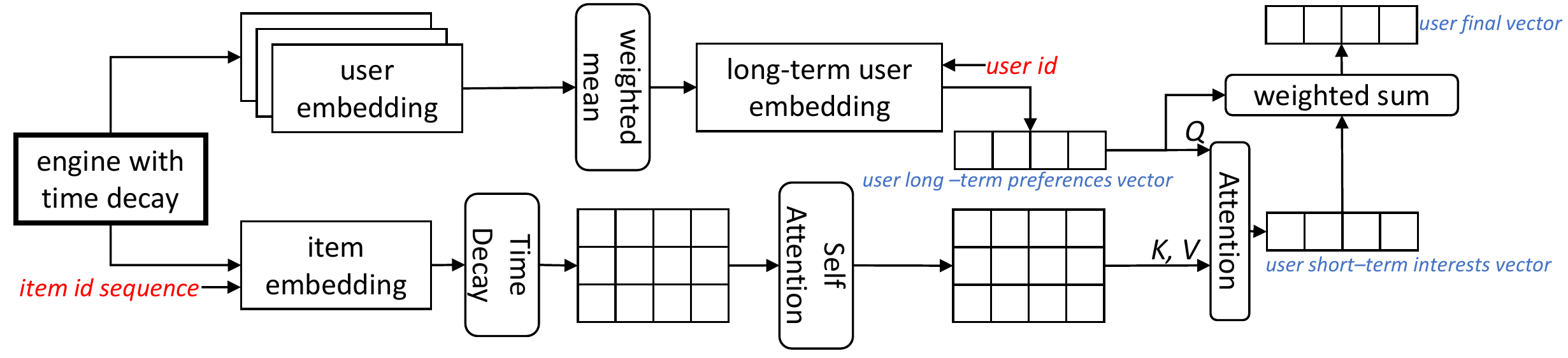}
  \caption{Personalized dynamic interaction pattern modeling module.}
  \label{fig.attention}
\end{figure}

Both Offline module and Online module cannot model the evolution trends of the dynamic interaction graph due to lacking the ability of memorization.
The incremental graph embedding engine captures collaborative relationships shared among similar users, but the interaction pattern of each user in his/her interaction sequence is irrelevant to his/her preference, i.e., interaction patterns are unique instead of collaborative with like-minded users.
Therefore, we propose an additional downstream module to capture user personalized dynamic interaction patterns as illustrated in Figure~\ref{fig.attention}.

\subsection{Dynamic Time Decay}
\label{sec.decay}
We define the process from the execution of an Offline module to the execution of the next Offline module a \textit{stage}.
At the beginning of the $i$-th stage, when constructing the historical interaction matrix, we decay the historical interaction score as $\exp\{\beta(t/{T_i}-1)\}$, where $T_i$ is the beginning time of the $i$-th stage and $\beta>0$ is the decay coefficient.
This implies that, for each stage, the score of historical interaction $(t<T_i)$ is less than $1$, while the score of interactions in current stage $(t\geq T_i)$ is greater than or equal to 1.
The decay should be memoryless, i.e., the decay factor is unchanged for the same time difference {(more
discussion can be found in the Appendix \ref{app.decay})}.
This requires that ${\beta}/{T}$ is a constant. 
Thus, for stage $i$ and stage $j$, their decay coefficients have the following relationship: ${\beta_i}/{\beta_j}={T_i}/{T_j}$.

Time decay makes the model more relevant to recent interactions through selective forgetting.
It can help the model capture the interaction pattern that users often interact with the recently interacted items in many link prediction tasks.
However, it cannot automatically capture the personalized dynamic interaction pattern of each user. Thus, we propose to combine attention mechanism with dynamic time decay to achieve personalized interaction pattern modelling.

\subsection{Attention Module}
{
For a user, we use the weighted average of $a$ recent user embeddings as his/her long-term embedding:
\begin{equation}
\textstyle\mathbf{e}_{long}=\sum_{r=1}^a\frac{1}{r}\mathbf{e}^{(r)}_u,
\end{equation}
where $\mathbf{e}^{(r)}_u\in\mathbb{R}^k$ is the corresponding user embedding, that is, a row of the $E_U$, at the $r$-th time step.
Then, we concatenate the decayed embeddings of the user's recent $b$ items as $S_u\in\mathbb{R}^{k\times b}$ and apply self-attention and attention to obtain the short term preference embedding $\mathbf{e}_{short}$ as follows:
\begin{equation}
    S_u=\mathbf{e}_i^{(1)}||\cdots||\mathbf{e}_i^{(b)},~~
    S_u'=\frac{S_u^\top S_u}{\sqrt{k}}S_u^\top,~~
    \mathbf{e}_{short}=(S_u')^\top\frac{S_u'\mathbf{e}_{long}}{\sqrt{k}}.
\end{equation}
Finally, we obtain the final user embedding by a weighted average:
\begin{equation}
\mathbf{e}=\lambda\mathbf{e}_{short}+(1-\lambda)\mathbf{e}_{long}.
\end{equation}
We use the dot product between current item embeddings and user final embedding: $E_I\cdot\mathbf{e}$ as the scores for predicting the link between the user and all items.
It should be noted that both self-attention and attention operations in our design require no weight matrices, i.e., they are parameter-free.
The whole process is shown in the Figure~\ref{fig.attention}.
}
\paragraph{Discussion}
A user has his/her own historical interactions, corresponding to personalization, and the recent historical interactions change over time, corresponding to temporal dynamic.
A user's long-term preference vector should be changing smoothly, which is in line with the characteristics of long-term interests.
A user's short-term preference vector is a linear combination of the embeddings of recently interacted items, which changes rapidly over time and is in line with the characteristics of short-term interests.
{Time decay of item sequences plays a key role in position embedding, which indicates that we will pay more attention to the recently interacting items.}
Self-attention can integrate the information of the whole sequence, and attention makes items with high similarity to users more likely to be interacted again.
Compared with time decay, the attention module is data-driven and can automatically adapt to the personalized and dynamic interaction patterns by making full use of the recent interactions.
However, compared with time decay, attention may be disturbed by many unnecessary patterns. Our empirical studies confirm that time decay and attention are complementary to each other and should be combined in the model.

\section{Experiments}
\paragraph{Datasets}
For the future item recommendation task, we use the Amazon Video, Amazon Game~\cite{he2016ups},  MovieLens-1M (ML-1M) and MovieLens-100K (ML-100K)~\cite{harper2015movielens}.
For the next interaction prediction task, we use Wikipedia and LastFM~\cite{kumar2019predicting}.
For all datasets, we use the first 80\% interactions as training set,
the following 10\% interactions as validation set, and the last 10\% interactions as test set.

\paragraph{Metrics}
For the future item recommendation task, we use \textit{Recall@10} to evaluate models.
For the next interaction prediction task, we use \textit{MRR} and \textit{Hit@10} to evaluate models.
For all experiments, we report the results on the test set when the models achieve the optimal results on  verification sets.

\paragraph{Baselines}
For the future item recommendation task, we compare with LightGCN~\cite{he2020lightgcn}, Time-LSTM~\cite{zhu2017next}, RRN~\cite{wu2017recurrent}, DeepCoevolve~\cite{dai2016deep}, JODIE~\cite{kumar2019predicting} and CoPE~\cite{zhang2021cope}.
For the next interaction prediction task, we compare with Time-LSTM~\cite{zhu2017next}, RRN~\cite{wu2017recurrent}, LatentCross~\cite{beutel2018latent}, CTDNE~\cite{nguyen2018continuous}, DeepCoevolve~\cite{dai2016deep}, JODIE~\cite{kumar2019predicting} and CoPE~\cite{zhang2021cope}.
Among all baselines,  LightGCN~\cite{he2020lightgcn} is the only static graph representation learning method for recommendation tasks.
{These baselines include TPP-based, RNN-based and GNN-based methods. For clarity, we discuss and compare these methods with \ours in the Appendix \ref{app.diff}.}

\subsection{Future item recommendation}
We use this task to verify whether the model can accurately predict user future interactions according to their historical interactions, which is a typical application of dynamic interaction graphs in recommender system. 
In this task, a user interacts with an item only once at most.

The results are shown in Table~\ref{table.item}, in which the results of baselines are reported by CoPE~\cite{zhang2021cope}.
To be fair, we do not allow JODIE, CoPE and \ours to update models during test phase (marked with \textit{*}).
In addition, since all baselines do not integrate attribute, we provide both the results of \ours with (\textit{with-attr}) and without (\textit{no-attr}) attributes.
It can be observed that \ours*(no-attr) achieves better accuracy than all baselines on all datasets.
Compared with truncated SVD, \ours*(no-attr) only adds the proposed  frequency-aware reconstruction module and the personalized interaction pattern modeller.
Further, \ours*(with-attr) outperforms \ours*(no-attr) due to the proposed attribute-integrated SVD, which verifies the effectiveness of this module.
Among all baselines, LightGCN performs the worst, because it is the only static GNN model which cannot capture the dynamic characteristics of the interaction graph.

\begin{table}\scriptsize
\caption{Accuracy comparison with state-of-the-art methods on two link prediction tasks.}
\centering
\subtable[Future item recommendation]{  
\begin{tabular}{c|cccc}
\hline
                      & Video                         & Game                          & ML-100K                       & ML-1M                         \\
\multicolumn{1}{c|}{} & \multicolumn{1}{c}{Recall} & \multicolumn{1}{c}{Recall} & \multicolumn{1}{c}{Recall} & \multicolumn{1}{c}{Recall} \\ \hline
LightGCN              & 0.036                         & 0.026                         & 0.025                         & 0.029                         \\
Time-LSTM             & 0.044                         & 0.020                         & 0.058                         & 0.033                         \\
RRN                   & 0.068                         & 0.029                         & 0.065                         & 0.043                         \\
DeepCoevolve          & 0.050                         & 0.027                         & 0.069                         & 0.030                         \\
JODIE*                & 0.078                         & 0.035                         & 0.074                         & 0.035                         \\
CoPE*                 & {\ul 0.088}                   & {\ul 0.047}                   & {\ul 0.081}                   & {\ul 0.049}                   \\
{\bf\ours*(no attr)}        & \textbf{0.113}                & \textbf{0.050}                & \textbf{0.114}                & \textbf{0.053}                \\ \hline
{\bf\ours*(with attr)}      & -                             & -                             & \textbf{0.149}                & \textbf{0.065}                \\ \hline
\end{tabular}
\label{table.item}  
}  
\quad
\subtable[Next interaction prediction]{ 
\begin{tabular}{c|cc|cc}
\hline
             & \multicolumn{2}{c|}{Wikipedia}  & \multicolumn{2}{c}{LastFM}      \\
             & MRR            & Hit         & MRR            & Hit         \\ \hline
Time-LSTM    & 0.247          & 0.342          & 0.068          & 0.137          \\
RRN          & 0.522          & 0.617          & 0.089          & 0.182          \\
LatentCross  & 0.424          & 0.481          & 0.148          & 0.227          \\
CTDNE        & 0.035          & 0.056          & 0.010          & 0.010          \\
DeepCoevolve & 0.515          & 0.563          & 0.019          & 0.039          \\
JODIE        & 0.746          & 0.822          & {\ul 0.195}    & 0.307          \\
CoPE         & {\ul 0.750}    & \textbf{0.890} & \textbf{0.200} & {\ul 0.446}    \\
{\bf\ours}         & \textbf{0.786} & {\ul 0.852}    & {\ul 0.195}          & \textbf{0.453} \\ \hline
\end{tabular}
\label{table.interaciton}
}  
\end{table}

\subsection{Next interaction prediction}
\label{sec.interaction}

We use this task to verify whether the model can accurately predict a user's next interaction according to the user's historical interactions up to the current timestamp, which is a kind of user behavior prediction problem.
In this task, a user may interact with an item for many times.

The results are shown in Table~\ref{table.interaciton}, in which the results of baselines are also reported by CoPE~\cite{zhang2021cope}. There are no user/item attributes in Wikipedia and LastFm, so we use all proposed modules except the  attribute-integrated SVD in \ours.
Since JODIE, CoPE and \ours can update models in test time, they significantly outperform the other methods without test time training.
\ours can outperform JODIE and achieve comparable accuracy with CoPE.
Although the accuracy of \ours has no obvious advantage over CoPE in this task, we will show later that \ours can significantly outperform CoPE in computation efficiency due to no learnable parameters.

\subsection{Ablation Studies}

\begin{table}[]\scriptsize
\caption{The ablation studies on the future item recommendation task.}
\label{table.ablation.item}
\centering
\begin{tabular}{c|cc|cccc}
\hline
     & Pattern Modeller & Matrix Reconstruction & Video & Game  & ML-100K & ML-1M \\ \hline
A    & \xmark     & \xmark                         & 0.073 & 0.023 & 0.051   & 0.045 \\
B    & \xmark     & \cmark                         & 0.107 & 0.041 & 0.074   & 0.051 \\
C    & \cmark     & \xmark                         & 0.080 & 0.025 & 0.113   & 0.052 \\
{\bf\ours*(no attr)} & \cmark     & \cmark                         & 0.113 & 0.050 & 0.114   & 0.053 \\ \hline
\end{tabular}
\end{table}

\paragraph{Future item recommendation}
We use A, B, C to refer to ablative variants of \ours, and check or cross marks indicate whether the corresponding module exists.
As shown in Table~\ref{table.ablation.item}, when the personalized interaction pattern
modeller or frequency-aware reconstruction module is not adopted, the results are suboptimal, which confirms the effectiveness of these two modules.

\paragraph{Next interaction prediction}
We use D-I to refer to ablative variants of \ours, and check or cross marks indicate whether the corresponding module exists.
In addition, we use an intuitively effective baseline, called \textit{Last-k}, which takes the recent $k$ items that users interacted with as predictions.
The results are shown in Table~\ref{table.ablation.interaction}.
Note that the result of \textit{Last-1} is Hit@1, {so we marked its result with *}.
D dose not update in test phase;
E only uses offline SVD to update guided by Monitor in test phase;
F updates in real time, but without offline restart.
G is better than D, E and F, which confirms the effectiveness of the Online-Monitor-Offline architecture.
\ours is better than H and I, indicating the effectiveness of both dynamic time decay module and attention module.
Dynamic time decay is very effective on Wikipedia, while attention has stronger effects on LastFm.
This is due to the phenomenon that many users continuously interact with the same item in Wikipedia, which can be proved by the excellent results of \textit{Last-k} on Wikipedia.
{It should be emphasized that, we think the upstream incremental graph embedding engine and the downstream personalized dynamic interaction pattern modeller are independent, and different implementations of the downstream modules can make our model deal with different downstream tasks. Therefore, during the ablation study, we do not perform cross-component ablation studies between the upstream engine and the downstream modeller, but only perform ablation studies within each component (i.e., within the upstream engine and within the downstream modeller).}

\begin{figure}[t]
  \centering
  \begin{minipage}[t]{0.6\linewidth}
    \centering
    \includegraphics[width=1.\linewidth]{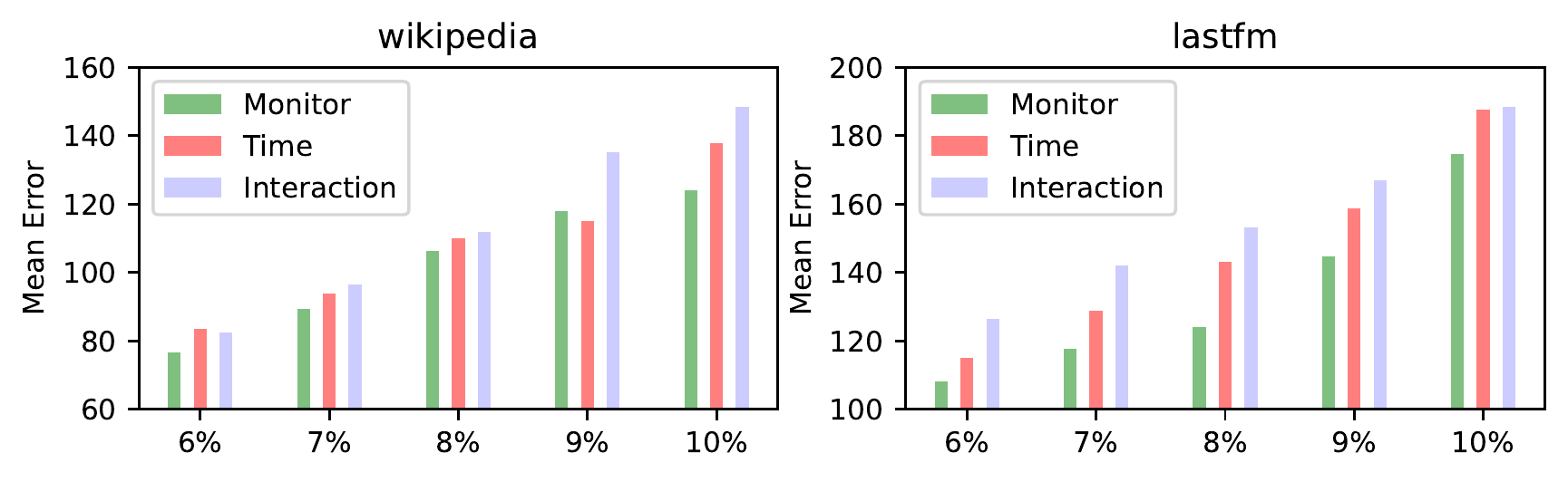}
    \caption{Performance comparison of three restart methods.}
    \label{fig.monitor}
  \end{minipage}
  \begin{minipage}[t]{0.35\linewidth}
    \centering
    \includegraphics[width=1.\linewidth]{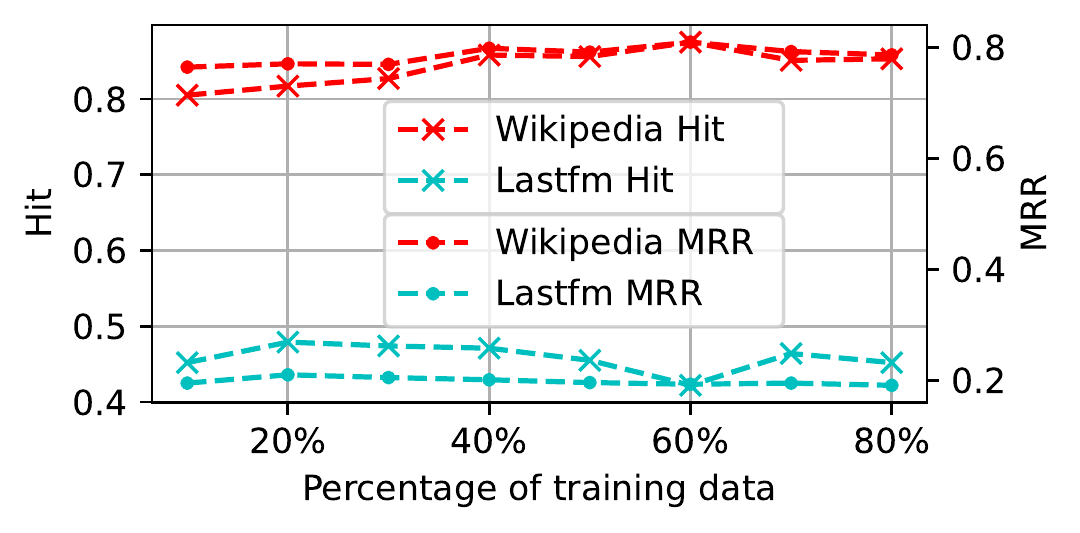}
    \caption{Robustness.}
    \label{fig.robustness}
  \end{minipage}
\end{figure}

\paragraph{Monitor}
We compare the average approximation errors of Monitor, \textit{Interaction} (restart with fixed number of interactions) and \textit{Time} (restart with fixed time), when the number of restarts are the same.
First, we execute online SVD to obtain the total approximation error, take 6\% - 10\% of the total error as the threshold of Monitor, and finally run Time and Interaction according to the number of restarts of Monitor.
The results are shown in Figure \ref{fig.monitor}.
Compared with the average approximation errors of Monitor, the errors of Interaction and Time are 10.93\% and 5.28\% higher on Wikipedia and 16.80\% and 9.67\% higher on LastFm, respectively, indicating that Monitor is more effective.

\begin{table}[]\scriptsize
\caption{The ablation studies on the next interaction prediction task.}
\label{table.ablation.interaction}
\centering
\begin{tabular}{c|cccc|cc|cc}
\hline
          & \multicolumn{4}{c|}{}          & \multicolumn{2}{c|}{Wikipedia} & \multicolumn{2}{c}{LastFM} \\
\textbf{} & Offline & Online & Decay & Attention & MRR            & Hit@10        & MRR          & Hit@10      \\ \hline
Last-1    & -       & -      & -     & -         & 0.775*         & 0.775*        & 0.098*       & 0.098*      \\
Last-10   & -       & -      & -     & -         & 0.792          & 0.842         & 0.139        & 0.263       \\ \hline
D         & \xmark  & \xmark & \xmark & \xmark   & 0.441          & 0.620         & 0.074        & 0.186       \\
E         & \cmark  & \xmark & \xmark & \xmark    & 0.510          & 0.703         & 0.089        & 0.222       \\
F         & \xmark  & \cmark & \xmark & \xmark    & 0.497          & 0.715         & 0.104        & 0.251       \\ 
G         & \cmark  & \cmark & \xmark & \xmark    & 0.530          & 0.739         & 0.11         & 0.256       \\ \hline
H         & \cmark  & \cmark & \cmark & \xmark    & 0.779          & 0.851         & 0.163        & 0.348       \\
I         & \cmark   & \cmark  & \xmark & \cmark  & 0.541          & 0.747         & 0.190        & 0.446     \\
{\bf\ours}      & \cmark   & \cmark  & \cmark & \cmark    & 0.786          & 0.852         & 0.195        & 0.453       \\ \hline
\end{tabular}
\end{table}

\subsection{Other Studies}
\paragraph{Running time}
\begin{table}[tbh!]\scriptsize
\caption{Total running time comparison. The times of speedup is shown in the parentheses.}
\label{table.running-time}
\centering
\begin{tabular}{ccc|ccc}
\hline
\multicolumn{3}{c|}{Wikipedia} &
\multicolumn{3}{c}{Lastfm} \\
\ours    & JODIE    & CoPE     & \ours  & JODIE    & CoPE  \\ \hline
9.7min        & 350.0min (36.1X)    & 3,589.1min (370.0X) & 54.8min       & 15,790.0min (288.1X) & 51,212.5min (934.5X) \\ \hline
\end{tabular}
\end{table}
We use the next interaction prediction task to study the efficiency of JODIE, CoPE and \ours.
Other baselines, such as Time-LSTM~\cite{zhu2017next}, RRN~\cite{wu2017recurrent}, 
LatentCross~\cite{beutel2018latent} and CTDNE~\cite{nguyen2018continuous} show comparable efficiency with JODIE~\cite{kumar2019predicting} and thus are omitted.
JODIE and CoPE both run 50 epochs and choose the best performing model on validation set as the optimal model, but \ours only requires multiple runs of offline/online SVD.
As shown in Table~\ref{table.running-time}, \ours is at least 36X faster than JODIE and at least 370X faster than CoPE, demonstrating its high efficiency. Besides, the running time of JODIE and CoPE increase significantly when the number of interactions increases, but the running time of \ours only increases moderately which is more desirable for real-world applications.

\paragraph{Cold-start}
\label{sec.cold}
We use the future item recommendation task to study the effects of attribute-integrated SVD in cold-start setting.
We define the users who have not appeared in training set and verification set as cold-start users.
For ML-1M, there are 29 cold-start users.
When attributes are not integrated, \ours randomly recommends items to these users.
For 17 of the cold-start users whose Recall@10 is not 0, \ours increases their average Recall@10 from 0.011 to 0.029, achieving a relative increase of \textbf{166\%}. 
For 7 cold-start users whose Recall@10 is 0, \ours increases their average Recall@10 to \textbf{0.100}, which is very high compared with the average Recall@10 of all users in ML-1M. 

\paragraph{Robustness}
For the next interaction prediction task, we change the proportion of the training set to verify the robustness of \ours in different levels of data sparsity.
We change the percentage of the training set from 10\% to 80\%, next 10\% interactions after the training set as the validation set, and next 10\% interactions after the validation set as the test set.
The results are shown in Figure~\ref{fig.robustness}, in which we can observe that the accuracy of \ours is almost unaffected. This experiment demonstrate that that \ours has strong robustness to the scale of training data.
The results of many baselines can also be found in JODIE~\cite{kumar2019predicting}.

\paragraph{Under represented groups}
{There could be recommendation bias concerns on under represented groups in specific applications and the existence of under represented groups usually comes from the biases during the data collection process.
Some bias mitigation techniques are specially designed to solve this problem~\cite{saxena2021exploring}.
We carry out further experiments on ML-100K.
First, we group the users by gender and find that there are 670 male users (71.0\%) and 273 female users (29.0\%) in this dataset.
Male users had 753,313 interaction records (74.4\%), and female users had 246,298 interaction records (25.6\%).
This shows that there is bias of gender distribution in this dataset.
Then, we remove the attribute-integrated SVD module from our method and calculate the performance of our model on male and female users. We find that without using attributes, the Recall of male users is 0.124 and Recall of female users is 0.085. Male users are with 45.9\% higher Recall than female users.
Finally, we take the attribute-integrated SVD module back to see how the performance differs. We find that after using attributes, the Recall of male users increases to 0.156, Recall of female users increases to 0.127. Male users are with 22.8\% higher Recall than female users.
To sum up, due to the bias of gender distribution in the ML-100K dataset, there is a clear bias in recommendation accuracy, i.e., male users are with 45.9\% higher Recall than female users even without using user attributes in the model. However, to our surprise, the bias in the recommendation results becomes less significant after using user attributes in our model, i.e., male users are with only 22.8\% higher Recall than female users when user attributes are incorporated.
We think the reason is that user attributes can help these under represented groups (female users) to improve their embedding quality.}

\section{Conclusion}
We propose \ours, a parameter-free dynamic graph embedding method for link prediction. 
By modelling collaborative relationships and personalized dynamic interaction patterns separately, \ours can alleviate the noises when the two key factors diverge, leading to higher accuracy. 
Specifically, we propose an incremental graph embedding engine for real-time dynamic graph embedding via a novel Online-Monitor-Offline architecture and a personalized dynamic interaction pattern modeller with dynamic time decay and attention. Since there are no learnable parameters, \ours can avoid the time-consuming back-propagation, leading to significantly higher efficiency.
One limitation of this work is that we only explore the link prediction task, leaving other downstream machine learning tasks, such as node classification, as future work.
{Another line of future work is to extend the high level design of our method to other kinds of parameterized models. Although we empirically find that our method can mitigate rating bias by utilizing user/item attributes, practitioners should pay more attention to the biases during data collection before using our method.}

\begin{ack}
This work was supported by the National Natural Science Foundation of China (NSFC) under Grants 61932007 and 62172106.
\end{ack}

\bibliographystyle{abbrv}  
\bibliography{neurips_2022}   

\begin{thebibliography}{10}

\bibitem{beutel2018latent}
A.~Beutel, P.~Covington, S.~Jain, C.~Xu, J.~Li, V.~Gatto, and E.~H. Chi.
\newblock Latent cross: Making use of context in recurrent recommender systems.
\newblock In {\em Proceedings of the Eleventh ACM International Conference on
  Web Search and Data Mining}, pages 46--54, 2018.

\bibitem{brand2006fast}
M.~Brand.
\newblock Fast low-rank modifications of the thin singular value decomposition.
\newblock {\em Linear algebra and its applications}, 415(1):20--30, 2006.

\bibitem{cao2021deep}
J.~Cao, X.~Lin, X.~Cong, S.~Guo, H.~Tang, T.~Liu, and B.~Wang.
\newblock Deep structural point process for learning temporal interaction
  networks.
\newblock In {\em Joint European Conference on Machine Learning and Knowledge
  Discovery in Databases}, pages 305--320. Springer, 2021.

\bibitem{chang2020continuous}
X.~Chang, X.~Liu, J.~Wen, S.~Li, Y.~Fang, L.~Song, and Y.~Qi.
\newblock Continuous-time dynamic graph learning via neural interaction
  processes.
\newblock In {\em Proceedings of the 29th ACM International Conference on
  Information \& Knowledge Management}, pages 145--154, 2020.

\bibitem{chen2021multi}
L.~Chen, S.~Yu, D.~Lyu, and D.~Wang.
\newblock Multi-relation aware temporal interaction network embedding.
\newblock {\em arXiv preprint arXiv:2110.04503}, 2021.

\bibitem{dai2016deep}
H.~Dai, Y.~Wang, R.~Trivedi, and L.~Song.
\newblock Deep coevolutionary network: Embedding user and item features for
  recommendation.
\newblock {\em arXiv preprint arXiv:1609.03675}, 2016.

\bibitem{goldberg1992using}
D.~Goldberg, D.~Nichols, B.~M. Oki, and D.~Terry.
\newblock Using collaborative filtering to weave an information tapestry.
\newblock {\em Communications of the ACM}, 35(12):61--70, 1992.

\bibitem{harper2015movielens}
F.~M. Harper and J.~A. Konstan.
\newblock The movielens datasets: History and context.
\newblock {\em Acm transactions on interactive intelligent systems (tiis)},
  5(4):1--19, 2015.

\bibitem{he2016ups}
R.~He and J.~McAuley.
\newblock Ups and downs: Modeling the visual evolution of fashion trends with
  one-class collaborative filtering.
\newblock In {\em proceedings of the 25th international conference on world
  wide web}, pages 507--517, 2016.

\bibitem{he2020lightgcn}
X.~He, K.~Deng, X.~Wang, Y.~Li, Y.~Zhang, and M.~Wang.
\newblock Lightgcn: Simplifying and powering graph convolution network for
  recommendation.
\newblock In {\em Proceedings of the 43rd International ACM SIGIR conference on
  research and development in Information Retrieval}, pages 639--648, 2020.

\bibitem{kang2021online}
H.~Kang, J.-H. Ho, D.~Mesquita, J.~P{\'e}rez, and A.~H. Souza.
\newblock Online graph nets.
\newblock 2021.

\bibitem{kazemi2020representation}
S.~M. Kazemi, R.~Goel, K.~Jain, I.~Kobyzev, A.~Sethi, P.~Forsyth, and
  P.~Poupart.
\newblock Representation learning for dynamic graphs: A survey.
\newblock {\em J. Mach. Learn. Res.}, 21(70):1--73, 2020.

\bibitem{kefato2021dynamic}
Z.~Kefato, S.~Girdzijauskas, N.~Sheikh, and A.~Montresor.
\newblock Dynamic embeddings for interaction prediction.
\newblock In {\em Proceedings of the Web Conference 2021}, pages 1609--1618,
  2021.

\bibitem{kumar2019predicting}
S.~Kumar, X.~Zhang, and J.~Leskovec.
\newblock Predicting dynamic embedding trajectory in temporal interaction
  networks.
\newblock In {\em Proceedings of the 25th ACM SIGKDD international conference
  on knowledge discovery \& data mining}, pages 1269--1278, 2019.

\bibitem{Li2017nips}
D.~Li, C.~Chen, W.~Liu, T.~Lu, N.~Gu, and S.~Chu.
\newblock Mixture-rank matrix approximation for collaborative filtering.
\newblock In {\em Advances in Neural Information Processing Systems}, pages
  477--485, 2017.

\bibitem{li2020dynamic}
X.~Li, M.~Zhang, S.~Wu, Z.~Liu, L.~Wang, and S.~Y. Philip.
\newblock Dynamic graph collaborative filtering.
\newblock In {\em 2020 IEEE International Conference on Data Mining (ICDM)},
  pages 322--331. IEEE, 2020.

\bibitem{nguyen2018continuous}
G.~H. Nguyen, J.~B. Lee, R.~A. Rossi, N.~K. Ahmed, E.~Koh, and S.~Kim.
\newblock Continuous-time dynamic network embeddings.
\newblock In {\em Companion Proceedings of the The Web Conference 2018}, pages
  969--976, 2018.

\bibitem{nt2019revisiting}
H.~Nt and T.~Maehara.
\newblock Revisiting graph neural networks: All we have is low-pass filters.
\newblock {\em arXiv preprint arXiv:1905.09550}, 2019.

\bibitem{sarwar2000application}
B.~Sarwar, G.~Karypis, J.~Konstan, and J.~Riedl.
\newblock Application of dimensionality reduction in recommender system-a case
  study.
\newblock Technical report, Minnesota Univ Minneapolis Dept of Computer
  Science, 2000.

\bibitem{sarwar2001item}
B.~Sarwar, G.~Karypis, J.~Konstan, and J.~Riedl.
\newblock Item-based collaborative filtering recommendation algorithms.
\newblock In {\em Proceedings of the 10th international conference on World
  Wide Web}, pages 285--295, 2001.

\bibitem{saxena2021exploring}
S.~Saxena and S.~Jain.
\newblock Exploring and mitigating gender bias in recommender systems with
  explicit feedback.
\newblock {\em arXiv preprint arXiv:2112.02530}, 2021.

\bibitem{shchur2019intensity}
O.~Shchur, M.~Bilo{\v{s}}, and S.~G{\"u}nnemann.
\newblock Intensity-free learning of temporal point processes.
\newblock {\em arXiv preprint arXiv:1909.12127}, 2019.

\bibitem{shen2021powerful}
Y.~Shen, Y.~Wu, Y.~Zhang, C.~Shan, J.~Zhang, B.~K. Letaief, and D.~Li.
\newblock How powerful is graph convolution for recommendation?
\newblock In {\em Proceedings of the 30th ACM International Conference on
  Information \& Knowledge Management}, pages 1619--1629, 2021.

\bibitem{steck2011item}
H.~Steck.
\newblock Item popularity and recommendation accuracy.
\newblock In {\em Proceedings of the fifth ACM conference on Recommender
  systems}, pages 125--132, 2011.

\bibitem{steck2019markov}
H.~Steck.
\newblock Markov random fields for collaborative filtering.
\newblock {\em Advances in Neural Information Processing Systems}, 32, 2019.

\bibitem{stewart1990matrix}
G.~W. Stewart.
\newblock Matrix perturbation theory.
\newblock 1990.

\bibitem{tian2021streaming}
S.~Tian, T.~Xiong, and L.~Shi.
\newblock Streaming dynamic graph neural networks for continuous-time temporal
  graph modeling.
\newblock In {\em 2021 IEEE International Conference on Data Mining (ICDM)},
  pages 1361--1366. IEEE, 2021.

\bibitem{trivedi2017know}
R.~Trivedi, H.~Dai, Y.~Wang, and L.~Song.
\newblock Know-evolve: Deep temporal reasoning for dynamic knowledge graphs.
\newblock In {\em international conference on machine learning}, pages
  3462--3471. PMLR, 2017.

\bibitem{wang2021tcl}
L.~Wang, X.~Chang, S.~Li, Y.~Chu, H.~Li, W.~Zhang, X.~He, L.~Song, J.~Zhou, and
  H.~Yang.
\newblock Tcl: Transformer-based dynamic graph modelling via contrastive
  learning.
\newblock {\em arXiv preprint arXiv:2105.07944}, 2021.

\bibitem{wang2016coevolutionary}
Y.~Wang, N.~Du, R.~Trivedi, and L.~Song.
\newblock Coevolutionary latent feature processes for continuous-time user-item
  interactions.
\newblock {\em Advances in neural information processing systems}, 29, 2016.

\bibitem{wen2022trend}
Z.~Wen and Y.~Fang.
\newblock Trend: Temporal event and node dynamics for graph representation
  learning.
\newblock In {\em Proceedings of the ACM Web Conference 2022}, pages
  1159--1169, 2022.

\bibitem{wu2017recurrent}
C.-Y. Wu, A.~Ahmed, A.~Beutel, A.~J. Smola, and H.~Jing.
\newblock Recurrent recommender networks.
\newblock In {\em Proceedings of the tenth ACM international conference on web
  search and data mining}, pages 495--503, 2017.

\bibitem{xia22www}
J.~Xia, D.~Li, H.~Gu, J.~Liu, T.~Lu, and N.~Gu.
\newblock Fire: Fast incremental recommendation with graph signal processing.
\newblock In {\em Proceedings of the ACM Web Conference 2022}, WWW '22, page
  2360–2369. ACM, 2022.

\bibitem{yang2022few}
C.~Yang, C.~Wang, Y.~Lu, X.~Gong, C.~Shi, W.~Wang, and X.~Zhang.
\newblock Few-shot link prediction in dynamic networks.
\newblock In {\em Proceedings of the Fifteenth ACM International Conference on
  Web Search and Data Mining}, pages 1245--1255, 2022.

\bibitem{yu2020graph}
W.~Yu and Z.~Qin.
\newblock Graph convolutional network for recommendation with low-pass
  collaborative filters.
\newblock In {\em International Conference on Machine Learning}, pages
  10936--10945. PMLR, 2020.

\bibitem{zeng2021knowledge}
Z.~Zeng, C.~Xiao, Y.~Yao, R.~Xie, Z.~Liu, F.~Lin, L.~Lin, and M.~Sun.
\newblock Knowledge transfer via pre-training for recommendation: A review and
  prospect.
\newblock {\em Frontiers in big Data}, page~4, 2021.

\bibitem{zhang2021cope}
Y.~Zhang, Y.~Xiong, D.~Li, C.~Shan, K.~Ren, and Y.~Zhu.
\newblock Cope: Modeling continuous propagation and evolution on interaction
  graph.
\newblock In {\em Proceedings of the 30th ACM International Conference on
  Information \& Knowledge Management}, pages 2627--2636, 2021.

\bibitem{zhang2018timers}
Z.~Zhang, P.~Cui, J.~Pei, X.~Wang, and W.~Zhu.
\newblock Timers: Error-bounded svd restart on dynamic networks.
\newblock In {\em Thirty-second AAAI conference on artificial intelligence},
  2018.

\bibitem{zhou2022tgl}
H.~Zhou, D.~Zheng, I.~Nisa, V.~Ioannidis, X.~Song, and G.~Karypis.
\newblock Tgl: A general framework for temporal gnn training on billion-scale
  graphs.
\newblock {\em arXiv preprint arXiv:2203.14883}, 2022.

\bibitem{zhu2017next}
Y.~Zhu, H.~Li, Y.~Liao, B.~Wang, Z.~Guan, H.~Liu, and D.~Cai.
\newblock What to do next: Modeling user behaviors by time-lstm.
\newblock In {\em IJCAI}, volume~17, pages 3602--3608, 2017.

\bibitem{zuo2018embedding}
Y.~Zuo, G.~Liu, H.~Lin, J.~Guo, X.~Hu, and J.~Wu.
\newblock Embedding temporal network via neighborhood formation.
\newblock In {\em Proceedings of the 24th ACM SIGKDD international conference
  on knowledge discovery \& data mining}, pages 2857--2866, 2018.

\end{thebibliography}


\clearpage

\section*{Checklist}

\begin{enumerate}

\item For all authors...
\begin{enumerate}
  \item Do the main claims made in the abstract and introduction accurately reflect the paper's contributions and scope?
    \answerYes{} 
  \item Did you describe the limitations of your work?
    \answerYes{} Please refer to the conclusion section.
  \item Did you discuss any potential negative societal impacts of your work?
    \answerNo{} To the best of our knowledge, this work does not have potential negative social impacts.
  \item Have you read the ethics review guidelines and ensured that your paper conforms to them?
    \answerYes{} 
\end{enumerate}

\item If you are including theoretical results...
\begin{enumerate}
  \item Did you state the full set of assumptions of all theoretical results?
    \answerNA{}
        \item Did you include complete proofs of all theoretical results?
    \answerNA{}
\end{enumerate}

\item If you ran experiments...
\begin{enumerate}
  \item Did you include the code, data, and instructions needed to reproduce the main experimental results (either in the supplemental material or as a URL)?
    \answerNo{} We will release our source code and detailed instructions for reproducing our results upon acceptance of this paper.
  \item Did you specify all the training details (e.g., data splits, hyperparameters, how they were chosen)?
    \answerYes{} Dataset splits are described in the main paper. Hyperparameters are introduced in the Appendix.
        \item Did you report error bars (e.g., with respect to the random seed after running experiments multiple times)?
    \answerNo{} Since we split the datasets by time, there are not randomness in terms of training, validation and test sets, which is the same as many prior works.
        \item Did you include the total amount of compute and the type of resources used (e.g., type of GPUs, internal cluster, or cloud provider)?
    \answerYes{} These are introduced in the Appendix.
\end{enumerate}

\item If you are using existing assets (e.g., code, data, models) or curating/releasing new assets...
\begin{enumerate}
  \item If your work uses existing assets, did you cite the creators?
    \answerYes{}
  \item Did you mention the license of the assets?
    \answerYes{} These are introduced in the Appendix.
  \item Did you include any new assets either in the supplemental material or as a URL?
    \answerNo{}
  \item Did you discuss whether and how consent was obtained from people whose data you're using/curating?
    \answerYes{} These are introduced in the Appendix.
  \item Did you discuss whether the data you are using/curating contains personally identifiable information or offensive content?
    \answerYes{} These are introduced in the Appendix.
\end{enumerate}

\item If you used crowdsourcing or conducted research with human subjects...
\begin{enumerate}
  \item Did you include the full text of instructions given to participants and screenshots, if applicable?
    \answerNA{}
  \item Did you describe any potential participant risks, with links to Institutional Review Board (IRB) approvals, if applicable?
    \answerNA{}
  \item Did you include the estimated hourly wage paid to participants and the total amount spent on participant compensation?
    \answerNA{}
\end{enumerate}

\end{enumerate}


\clearpage
\appendix

\section{Discussion}
\subsection{{Connection between SVD and Frequency Analysis}}
\label{app.fre}
First, we introduce the concept of ``low frequency signals'' following the common practice in graph signal processing.

We use $R$ to represent the observed interaction matrix, $\tilde{R}$ to represent the user's real preference matrix, where  larger $\tilde{R}_{ij}$ indicates higher chance for user $i$ to interact with item $j$, and $\hat{R}$ to represent the predicted interaction matrix. We can use $S=R^\top R$ to represent the similarity between items, and $D$ to represent the degree matrix of $S$. The Laplace matrix of $S$ is defined as $L=D-S$.
We can take $\pmb{x}\in\mathbb{R}^n$ as a graph signal where each node is assigned with a scalar. The smoothness of the graph signal can be measured by the total variation defined as follows:
\begin{equation}
TV(\pmb{x})=\pmb{x}^TL\pmb{x}=\sum S_{ij}(x_i-x_j)^2.
\end{equation}
When the input graph signal is the real preference vector of user $u$, which means $\pmb{x}=\tilde{R}_{u}\in\mathbb{R}^n$: 1) if two items $i$ and $j$ $(i\not=j)$ are similar, i.e., $S_{ij}$ is large, the user's preferences for these two items will be similar, which means that $(\tilde{R}_{ui}-\tilde{R}_{uj})^2$ should be small and will not lead to excessive $TV(\pmb{x})$; 2) if two items $i$ and $j$ $(i\not=j)$ are dissimilar, i.e., $S_{ij}$ is small, the user often has different preferences for the two items, which means that $(\tilde{R}_{ui}-\tilde{R}_{uj})^2$ should be large, which also does not cause $TV(\pmb{x})$ to be too large because their similarity $S_{ij}$ is small.
In conclusion, if the real preference signal is used as input, the total variation should have a small value. However, due to the exposure noise and quantization noise in the observed interaction matrix~\cite{yu2020graph}, the total variation becomes larger when the input signal is the observed user interaction signal, which means $\pmb{x}=R_u$.
Therefore, the key to predict the real preference matrix through the observed interaction matrix is to design a low-pass filter to remove the high-frequency part of the observed interaction matrix.

Then, we explain why the reconstruction matrix obtained by truncated SVD is low-frequency, which is also related to graph signal processing.

The energy of the graph signal is defined as $E(\pmb{x})=||\pmb{x}||^2$. The normalized total variation of $\pmb{x}$ can be calculated with the Rayleigh quotient as 
\begin{equation}
Ray(\pmb{x})=\frac{TV(\pmb{x})}{E(\pmb{x})}=\frac{\pmb{x}^TL\pmb{x}}{\pmb{x}^T\pmb{x}}=\frac{\sum S_{ij}(x_i-x_j)^2}{\sum x_i^2}.
\end{equation}
As $L$ is real and symmetric, its eigendecomposition is given by $L=U\Lambda U^T$ where $\Lambda=diag(\lambda_1,\lambda_2,...,\lambda_n),\lambda_1\le \lambda_2\le...\le \lambda_n$, and $U=(\pmb{u}_1,\pmb{u}_2,...,\pmb{u}_n)$ with $\pmb{u}_i\in\mathbb{R}^{n}$ being the eigenvector for eigenvalue $\lambda_i$. We call $\tilde{\pmb{x}}=U^T\pmb{x}$ as the graph Fourier transform of the graph signal $\pmb{x}$ and its inverse transform is given by $\pmb{x}=U\tilde{\pmb{x}}$. Rayleigh quotient can be transformed into spectral domain as
\begin{equation}
Ray(\pmb{x})=\frac{\pmb{x}^TL\pmb{x}}{\pmb{x}^T\pmb{x}}=\frac{\pmb{x}^TU\Lambda U^T\pmb{x}}{\pmb{x}^TUU^T\pmb{x}}=\frac{\tilde{\pmb{x}}^T\Lambda \tilde{\pmb{x}}}{\tilde{\pmb{x}}^T\tilde{\pmb{x}}}=\frac{\sum\lambda_i\tilde{x}^2_i}{\sum\tilde{x}^2_i}.
\end{equation}
Take $\pmb{x}=\pmb{u}_i$, we can get $Ray(\pmb{u}_i)=\lambda_i$, indicating that the eigenvector corresponding to the small eigenvalue is smoother.

There are similar conclusions when we take $S=RR^\top$. SVD extends the signal on the node from scalar to vector. The three matrices obtained by truncated SVD correspond to the first $k$  eigenvectors of $RR^\top$, the first $k$ eigenvalues of $RR^\top$ and the first $k$ eigenvectors of $R^\top R$ respectively. Therefore, it only retains the eigenvectors with low frequency to reconstruct the interaction matrix, so its essence is an ideal low-pass filter. And their frequency is related to the magnitude of eigenvalues.

Recently, Nt et al.~\cite{nt2019revisiting} show that the method of graph neural network is essentially a low-pass graph filter. Shen et al.~\cite{shen2021powerful} show that matrix factorization methods, linear auto-encoder methods, and neighborhood-based methods can be equivalently described by designing  different forms of graph filters in graph signal processing. In addition, the method based on matrix factorization is proved to be equivalent to an infinite layer graph neural network.

\subsection{{Why $\pi/3$ is the Boundary?}}
\label{app.pi}
In summary, $\pi/3$ is the threshold to determine the effectiveness of model updates. Online updating towards angles greater than $\pi/3$ indicates the online updating is even worse than no updating at all. More detailed discussion is presented below.

\begin{figure}[h!]
  \centering
  \includegraphics[width=0.5\linewidth]{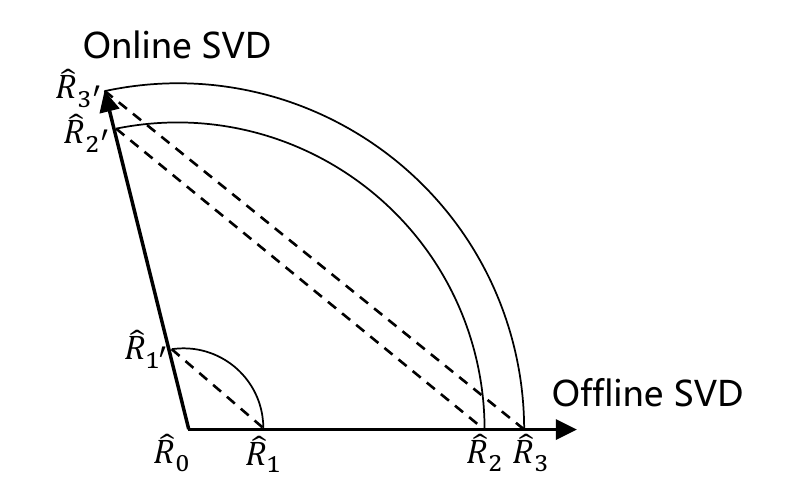}
  \caption{The evolution direction of Offline module and Online module is greater than $\pi/3$}
  \label{fig.monitor.greater}
\end{figure}

When we use offline SVD during the processes of $\hat{R}_0$->$\hat{R}_1$, $\hat{R}_1$->$\hat{R}_2$ and $\hat{R}2$->$\hat{R}_3$, there will be no approximation error. However, the online SVD used in the processes of $\hat{R}_0$->$\hat{R}_{1'}$, $\hat{R}_{1'}$->$\hat{R}_{2'}$ and $\hat{R}_{2'}$->$\hat{R}_{3'}$ has approximation errors. Thus, their evolution directions are not consistent. We use the F-norm of the difference between $\hat{R}_i$ and $\hat{R}_{i'}$ $(i=1,2,3)$ to measure the online approximation error. We cannot directly calculate this value in most cases, because the model will only execute the Online module in most cases. From Figure \ref{fig.monitor.motivation}, we find that the F-norm of the difference between $\hat{R}_0$ and $\hat{R}_{i'}$ (defined as \textit{distance} in this paper) is positively correlated with the F-norm of the difference between $\hat{R}_i$ and $\hat{R}_{i'}$ $(i=1,2,3)$, and we have verified this empirically in Appendix \ref{ap:monitor}. Thus, we use \textit{distance} to estimate the \textit{error}. As the approximation of the Offline module, the approximated value calculated by the Online module should not be worse than the result without updating, otherwise it means that the Online module is invalid. In this case, the approximation error of the reconstructed matrix $(\hat{R}_{1'}, \hat{R}_{2'}, \hat{R}_{3'})$ obtained by the Online module is even greater than the F-norm between the original matrix ($\hat{R}_0$) and the reconstructed matrix $(\hat{R}_1,\hat{R}_2,\hat{R}_3)$ obtained by the Offline module.
Figure \ref{fig.monitor.greater} shows the case when the evolution direction of Offline module and Online module is greater than $\pi/3$.
It can be seen that the F-norm of the difference between $\hat{R}_0$ and $\hat{R}_i$ is smaller than the F-norm of the difference between $\hat{R}_i$ and $\hat{R}_{i'}$ $(i=1,2,3)$. In this case, instead of using $\hat{R}_{i'}$ as the approximation of $\hat{R}_i$, it is better to directly use $\hat{R}_0$ as its approximation, i.e., online updating is even worse than no updating at all.

\subsection{The Positive Correlation Between Approximation Error and Distance}
\label{ap:monitor}
\begin{figure}[h!]
  \centering
  \includegraphics[width=0.8\linewidth]{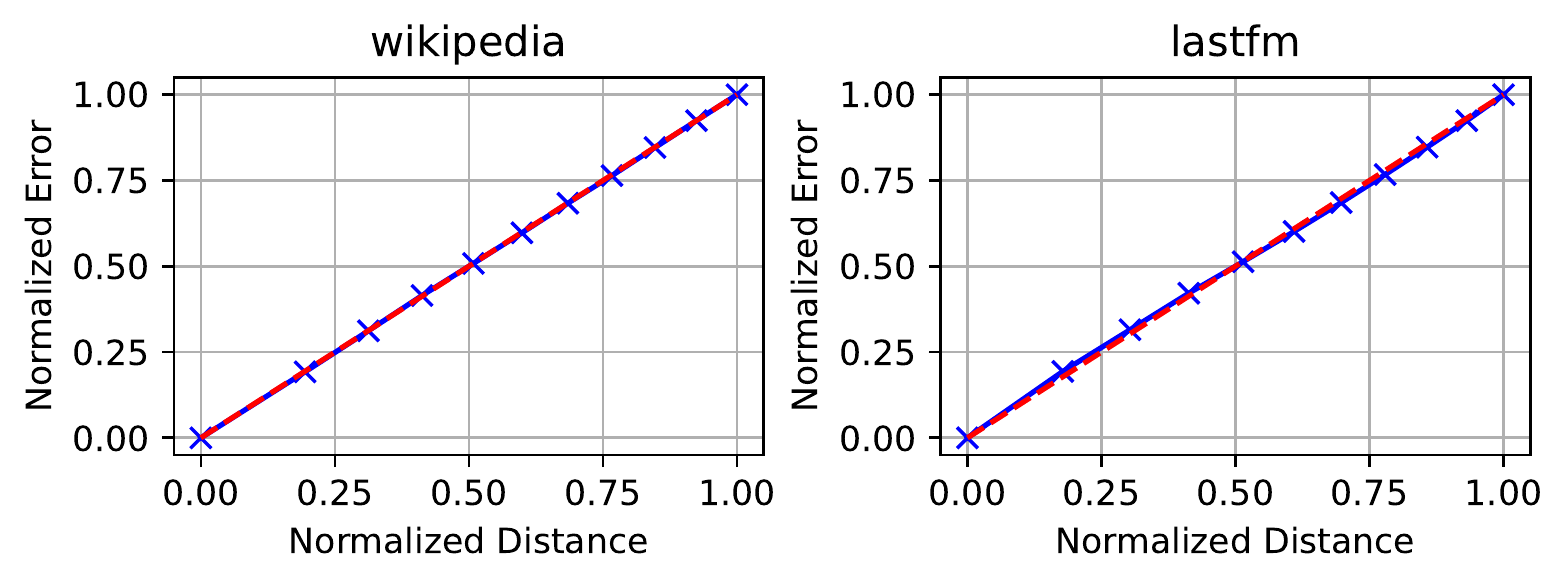}
  \caption{There is a strong positive correlation between \textit{distance} (the F-norm distance between the reconstructed matrix by online SVD at the current time step and the initial reconstructed matrix) and online approximation error.}
  \label{fig.monitor.curve}
\end{figure}

We first divide each dataset into $T=100$ time intervals according to the number of interactions, and run offline SVD as ground truth at the end of each time interval.
We start to execute the online SVD at the end of $t$-th time interval, and calculate the error of the online SVD and the \textit{distance} at the end of $t+\Delta$-th time interval.
The value of $\Delta$ is from 1 to 10, and the corresponding value of $t$ is from $1$ to $T-\Delta$.
Each $\Delta$ corresponds to a group of errors and a group of \textit{distance}.
By grouping according to $\Delta$, we can calculate the mean value of each group of errors and the mean value of \textit{distance}, respectively, and draw the curve to understand their relationship.
For $\Delta=1,...,10$, we define $d_\Delta$ and $e_\Delta$ as the corresponding mean of \textit{distance} and mean of online approximation errors. Then, we normalize $d_\Delta$ and $e_\Delta$ as the $x$-axis and $y$-axis, respectively, as follows:
\begin{align}
x_\Delta&=||d_\Delta||=\frac{d_\Delta^m}{max(d_1^m, ..., d_\Delta^m)}, \\ 
y_\Delta&=||e_\Delta||=\frac{e_\Delta}{max(e_1, ..., e_\Delta)}.
\end{align}
As shown in Figure \ref{fig.monitor.curve}, there is a positive correlation between the normalized error and the normalized \textit{distance}, indicating that it is reasonable to estimate the online approximation error using the \textit{distance} measure. Although the power $m$ varies between the two datasets, we can always find an appropriate $m$ so that we can fit the relationship between $d_\Delta$ and $e_\Delta$ almost by a straight line.

\subsection{{Memoryless Property of Time Decay Function}}
\label{app.decay}
The Online-Monitor-Offline architecture indicates that our model has the concept of stage.
We believe that the decay function should have no memories, that is, the decay ratio of the same time interval should be treated consistently in different stages. The reasons are explained in the following discussion.

Suppose there are four timestamps $t_1$, $t_2$, $t_3$, $t_4$, and $t_2-t_1=t_4-t_3$.
The decay function we used $f_i(t)=exp\{\beta_i(t/T_i -1)\}$ is memoryless. When these four timestamps are all in the same stage (e.g., $i$-th stage), we have $f_i(t_2)/f_i(t_1)=f_i(t_4)/f_i(t_3)$ using the decay function. When these four timestamps are in different stages, (e.g., $t_1$ and $t_2$ are at i-th stage, and $t_3$ and $t_4$ are at j-th stage ($i\neq j$)), we have $f_i(t_2)/f_i(t_1)=f_j(t_4)/f_j(t_3)$, when $\beta_i/\beta_j=T_i/T_j$.

We point out here that the linear decay function $g_i(t)=\beta_it/T_i$ is not memoryless. This is because we cannot ensure $g_i(t_2)/g_i(t_1)=g_j(t_4)/g_j(t_3)$ both in the above two cases.

\subsection{{Comparison between Different Models}}
\label{app.diff}
In order to more clearly explain the difference between \ours and TPP-based, RNN-based and GNN-based methods, we summarized this and presented the results in the Table \ref{table.comp}.

In all methods, only \ours is parameter-free.
Various dynamic graph learning methods use time decay mechanism, but in different forms.
For the methods based on temporal point process~\cite{trivedi2017know,wang2016coevolutionary,zuo2018embedding}, the time decay is reflected in the intensity function.
When RNN-based method~\cite{kumar2019predicting} predicts the user/item embeddings, it specifically considers the current user's previous interactions using RNN.
GNN-based method~\cite{zhang2021cope} adopts a neural ordinary differential equation to model the temporal dynamics of user/item embeddings.
It should be noted that the dynamic time decay in other methods can also be applied to our framework but may introduce learnable-parameters and thus hurt the computational efficiency.

TPP-based models~\cite{trivedi2017know,wang2016coevolutionary,zuo2018embedding} make use of collaboration through interaction between users and items.
RNN-based model~\cite{kumar2019predicting} uses a pair of coupled RNNs to model users and items respectively to make use of collaborative relationships. 
Because GNN naturally contains graph information, the model based on GNN~\cite{zhang2021cope} naturally uses the collaborative relationship through adjacency matrix.
Similar to the GNN-based model, the SVD method used by \ours also contains graph information, so collaborative information is naturally adopted.

\begin{table}[h!]\small
\caption{Comparison between different kinds of methods.}
\label{table.comp}
\centering
\begin{tabular}{c|c|c|c|c}
\hline
                                                                                & TPP-based                                                    & RNN-based   & GNN-based                                                                       & FreeGEM                                                      \\ \hline
parameter-free                                                                  & \xmark                                                            & \xmark           & \xmark                                                                               & \cmark                                                            \\ \hline
\begin{tabular}[c]{@{}c@{}}how to use\\ time information\end{tabular}           & \begin{tabular}[c]{@{}c@{}}intensity\\ function\end{tabular} & RNN         & \begin{tabular}[c]{@{}c@{}}neural ordinary\\ differential equation\end{tabular} & \begin{tabular}[c]{@{}c@{}}dynamic\\ time decay\end{tabular} \\ \hline
\begin{tabular}[c]{@{}c@{}}how to use\\ collaboration relationship\end{tabular} & \begin{tabular}[c]{@{}c@{}}interaction\\ events\end{tabular} & coupled RNN & GNN                                                                             & SVD                                                          \\ \hline
\end{tabular}
\end{table}

\section{Appendix}
\subsection{Statistics of the Datasets}
We use four publicly available datasets to evaluate the performance of \ours in the future item recommendation task, the detailed statistics of which are presented in Table~\ref{table.dataset.recommendation}. 
\begin{table}[h!]\small
\caption{Statistics of the datasets of the future item recommendation task.}
\label{table.dataset.recommendation}
\centering
\begin{tabular}{c|ccccc}
\hline
\textbf{Datasets} & \textbf{\# Users} & \textbf{\# Items} & \textbf{\# Interactions} & \textbf{\# Density} & \textbf{\# Unique Timestamps} \\ \hline
Amazon Video             & 5,130              & 1,685              & 37,126                    & 0.43\%              & 1,946                     \\
Amazon Game              & 24,303             & 10,672             & 231,780                   & 0.09\%              & 5,302                     \\
MovieLens-100K           & 943               & 1,349              & 99,287                    & 7.81\%              & 49,119                    \\
MovieLens-1M             & 6,040              & 3,416              & 999,611                   & 4.85\%              & 458,254   \\ \hline               
\end{tabular}
\end{table}

We use two publicly available datasets to evaluate the performance of \ours in the next interaction prediction task, the detailed statistics of which are presented in Table~\ref{table.dataset.interaction}. 
\begin{table}[h!]\small
\caption{Statistics of the datasets of the next interaction prediction task.}
\label{table.dataset.interaction}
\centering
\begin{tabular}{c|cccc}
\hline
\textbf{Dataset} & \textbf{\# Users} & \textbf{\# Items} & \textbf{\# Interactions} & \textbf{\# Unique Timestamps} \\ \hline
Wikipedia        & 8,227              & 1,000              & 157,474                   & 152,757                   \\
LastFM           & 980               & 1,000              & 1,293,103                  & 1,283,614                  \\ \hline
\end{tabular}
\end{table}

\subsection{Hyperparameter Settings}
Although there are no learnable parameter in \ours, we have several hyperparameters for model building and user preference fusion. In our experiments, we use a simple grid search method to obtain the optimal hyperparameters.
The hyperparameter search space of \ours in the future item recommendation task is presented in Table~\ref{table.hyper-search.recommendation}.

\begin{table}[h!]\small
\caption{Hyperparameter search space for future item recommendation task.}
\label{table.hyper-search.recommendation}
\centering
\begin{tabular}{ccccccc}
\hline
$\beta_1$ & $\alpha$ & $k_1$ & $k_2$, $k_3$, $k_4$, $k_5$ & $\alpha_1$ & $\alpha_2$ & $\alpha_3$ \\ \hline
1, ..., 100 & 2.0 & 1, 2, 4, 8, 16, 32, 64, 128, 256 & 0, 1 & 0, 1, 2 & 0, 1, 2 & 0, 1, 2 \\ \hline
\end{tabular}
\end{table}
After grid search, we find the optimal hyperparameters of \ours for the four datasets on the future item recommendation task in Table~\ref{table.hyper.recommendation}.
\begin{table}[h!]\small
\caption{Hyperparameter settings for future item recommendation task.}
\label{table.hyper.recommendation}
\centering
\begin{tabular}{c|ccccccccccccc}
\hline
        & $\beta_1$ & $\alpha$ & $k_1$ & $k_2$ & $k_3$ & $k_4$ & $k_5$ & $\alpha_1$ & $\alpha_2$ & $\alpha_3$ \\ \hline
Video   &  21.0   &  2.0     &   128 &   0   &  0    &  0    &  0        &   1        &    0       &    0      \\
Game    &  18.0   &  2.0     &   256 &   0   &  0    &  0    &  0        &   1        &    0       &    0     \\
ML-1M (no-attr)  &  60.0   &  2.0     &    8  &   0   &  0    &  0    &  0     &   1        &    0       &    0     \\
ML-100K (no-attr) &  60.0   &  2.0     &    1  &   0   &  0    &  0    &  0     &   1        &    0       &    0     \\ \hline
ML-1M (with-attr)   &  50.0   &  2.0     &    4  &   1   &  1    &  1    &  1      &   0        &    1       &    0    \\
ML-100K (with-attr) &  15.0   &  2.0     &    1  &   1   &  1    &  1    &  1     &   0        &    2       &    1   \\ \hline
\end{tabular}
\end{table}

The hyperparameter search space of \ours in the next interaction prediction task is presented in Table~\ref{table.hyper-search.interaction}.
\begin{table}[h!]\scriptsize
\caption{Hyperparameter search space for next interaction prediction task.}
\label{table.hyper-search.interaction}
\centering
\begin{tabular}{cccccccc}
\hline
          & $d$ & $\beta_1$ & $a$ & $b$ & $k_1$ & $\gamma$ & $\lambda$ \\ \hline
Wikipedia & 5, 10,..., 50 & 5, 10, ..., 50 & 1, 3, 5 & 1, 2, 3 & 128, 256, 512 & 1/2, 1/3, ..., 1/10 & 0.01, 0.02, ..., 1.00 \\
Lastfm    & 300, 400, ..., 800 & 1, 2, ..., 10 & 1, 3, 5 & 1, 2, 3 & 128, 256, 512 & 1/2, 1/3, ..., 1/10 & 0.01, 0.02, ..., 1.00 \\ \hline
\end{tabular}
\end{table}

After grid search, we find the optimal hyperparameters of \ours for the two datasets on the next interaction prediction task in Table~\ref{table.hyper.interaction}.
\begin{table}[h!]
\small
\caption{Hyperparameter settings for next interaction prediction task.}
\label{table.hyper.interaction}
\centering
\begin{tabular}{c|ccccccc}
\hline
          & $d$   & $\beta_1$ & $a$ & $b$ & $k_1$ & $\gamma$ & $\lambda$ \\ \hline
Wikipedia & 35.0  & 35.0      &  3  &  1  & 512   &  1/2     & 0.80 \\
LastFM    & 500.0 &  2.0      &  1  &  2  &  512  &  1/5     & 0.74 \\ \hline
\end{tabular}
\end{table}

{In addition, we have the following observations about the hyperparameters.}

{(1) Hyperparameters related to the attribute-integrated SVD module include $k_2, k_3, k_4, k_5$, and $\alpha_1, \alpha_2, \alpha_3$. Due to the inclusion of attribute information, this module can improve the prediction accuracy but also introduce several more hyperparameters. It can be observed in ablation experiments that our method can still surpass other methods even without using attribute information. Thus, in practice, attribute information can be used only for those users with scarce interactions or cold start users, which can significantly improve the accuracy as shown in Section \ref{sec.cold}. In addition, during hyperparameter search, we observe that using $k_1=k_2=k_3=k_4$ can achieve outstanding performance, even though they can be tuned separately.}

{(2) The hyperparameter that controls the restart of the Offline module has less significant effect on the prediction accuracy. Its role is to control the restart times according to the data scale. As we can see, in Wikipedia dataset with small data scale, we use the search interval of 5, 10, ..., 50, while in LastFm dataset with large data scale, we use the search interval of 300, 400, ..., 800.}

{(3) In the experiments, we find that for different datasets, the hyperparameter $\beta$, which controls the time decay is sensitive to the dataset. We can set higher priority for the searching of $\beta$. Luckily, $\beta$ is not very sensitive to the values of other hyperparameters, so that we can search other hyperparameters after the optimal $\beta$ is found.}

{(4) As shown in Table \ref{table.running-time}, the model training time of our method is much shorter compared to other methods for one group of hyperparameters, especially on larger dataset. Thus, we find that the overall running time (include hyperparameter searching) of our method is still much lower than the other methods.}

\subsection{Experimental Environment}
We run all the experiments on a server equipped with one NVIDIA TESLA T4 GPU and Intel(R) Xeon(R) Gold 5218R CPU. 
All the code of this work is implemented with Python 3.9.7.

\subsection{Copyrights of the Existing Assets}
All the code that we use to reproduce the results of the compared works is publicly available and permits usage for research purpose.

All the datasets that we use in the experiments are publicly available and permit usage for research purpose.

\end{document}